\documentclass[letterpaper]{article} % DO NOT CHANGE THIS
\usepackage{aaai2026} 

\usepackage{times}
\usepackage{helvet}
\usepackage{courier}
\usepackage[hyphens]{url}
\usepackage{graphicx}
\usepackage{caption}
\usepackage[dvipsnames,svgnames,x11names]{xcolor}
\usepackage{colortbl}
\usepackage{graphicx}
\usepackage{booktabs}
\usepackage{array}
\usepackage{makecell}
\usepackage{diagbox}

\usepackage{enumitem}
\usepackage[ruled,vlined]{algorithm2e}
\usepackage{natbib}
\usepackage{caption}
\usepackage[skip=3pt,font=small]{subcaption}
\usepackage{amsmath,amsfonts,mathtools}
\usepackage{multirow,tabularx}
\usepackage[capitalise,nameinlink]{cleveref}
\usepackage{xspace}
\usepackage{cuted}

\usepackage{acronym}
\usepackage[skip=3pt,font=small]{subcaption}
\usepackage{amsmath,amsfonts,mathtools}
\usepackage[dvipsnames,svgnames,x11names]{xcolor}
\usepackage{booktabs,tabularx,colortbl,multirow,array,makecell}
\usepackage[capitalise,nameinlink]{cleveref}
\usepackage{xspace}
\usepackage{cuted}

\usepackage{listings}
\DeclareCaptionStyle{ruled}{labelfont=normalfont,labelsep=colon,strut=off} % DO NOT CHANGE THIS
\definecolor{WeiColor}{rgb}{1,0.33,0.64}
\definecolor{myDarkBlue}{RGB}{55,81,139}
\definecolor{myLightBlue}{RGB}{158,186,217}
\definecolor{myLightGreen}{RGB}{126,171,85}
\definecolor{myRed}{RGB}{176,36,24}
\definecolor{QiColor}{RGB}{81,139,55}
\definecolor{myBlue}{RGB}{64,224,205}
\definecolor{red}{RGB}{255,0,0}
\definecolor{lightyellow}{RGB}{227,207,87}
\definecolor{lightred}{rgb}{0.5, 0.0, 0.13}
\definecolor{lightpurple}{rgb}{0.74, 0.2, 0.64}

\makeatletter
\DeclareRobustCommand\onedot{\futurelet\@let@token\@onedot}
\def\@onedot{\ifx\@let@token.\else.\null\fi}

\makeatother

\crefname{algorithm}{Alg.}{Algs.}
\Crefname{algocf}{Alg.}{Algs.}
\crefname{section}{Sec.}{Secs.}
\Crefname{section}{Section}{Sections}
\crefname{table}{Tab.}{Tabs.}
\Crefname{table}{Table}{Tables}
\crefname{figure}{Fig.}{Fig.}
\Crefname{figure}{Figure}{Figure}

\newcommand{\task}{\text{FloNa}\xspace}

\acrodef{flona}[\task]{\underline{Flo}or Plan Visual \underline{Na}vigation}
\acrodef{vint}[ViNT]{Visual Navigation Transformer}
\acrodef{nomad}[NoMaD]{Navigation with Goal Masked Diffusion}
\acrodef{mlp}[MLP]{Multi Layer Perceptron}
\acrodef{mse}[MSE]{mean squared error}

\usepackage{amsmath, amsfonts, multirow}
\usepackage{tcolorbox} % 加载 tcolorbox 包
\tcbuselibrary{theorems} % 使用 theorems 库以启用 \tcbhighmath 命令
\usepackage{csquotes}
\usepackage{float}
\makeatletter
\newcommand{\thickhline}{%
    \noalign {\ifnum 0=`}\fi \hrule height 1pt
    \futurelet \reserved@a \@xhline
}
\makeatother

\definecolor{DarkBlue}{RGB}{64,101,149}
\definecolor{azure}{rgb}{0.0, 0.5, 1.0}
\definecolor{gray}{rgb}{0.3, 0.3, 0.3}
\definecolor{DarkGreen}{RGB}{42,110,63}
\definecolor{myred}{RGB}{255,0,0} % 红色
\definecolor{mygray}{gray}{.9}
\usepackage{pifont}

\title{RGMP:  Recurrent Geometric-prior Multimodal Policy for Generalizable Humanoid Robot Manipulation}
\author{
    Xuetao Li\textsuperscript{\rm 1}\equalcontrib,
    Wenke Huang\textsuperscript{\rm 1}\equalcontrib,
    Nengyuan Pan\textsuperscript{\rm 2},
    Kaiyan Zhao\textsuperscript{\rm 1},
    Songhua Yang\textsuperscript{\rm 1},
    Yiming Wang\textsuperscript{\rm 3},
    Mengde Li\textsuperscript{\rm 4},
    Mang Ye\textsuperscript{\rm 1},
    Jifeng Xuan\textsuperscript{\rm 1},
    Miao Li\textsuperscript{\rm 1,4,5}\thanks{Corresponding author}
}
\affiliations{
    \textsuperscript{\rm 1}School of Computer Science, Wuhan University\\
    \textsuperscript{\rm 2}Faculty of Artificial Intelligence, Hubei University\\
    \textsuperscript{\rm 3}State Key Laboratory of Internet of Things for Smart City, University of Macau\\
    \textsuperscript{\rm 4}Institute of Technological Sciences, Wuhan University\\
    \textsuperscript{\rm 5}School of Robotics, Wuhan University\\
    \{xtli312, wenkehuang, yemang, jxuan, miao.li\}@whu.edu.cn
}

\usepackage{bibentry}
\begin{document}

\maketitle

\begin{abstract}
Humanoid robots exhibit significant potential in executing diverse human-level skills. However, current research predominantly relies on data-driven approaches that necessitate extensive training datasets to achieve robust multimodal decision-making capabilities and generalizable visuomotor control. These methods raise concerns due to the neglect of geometric reasoning in unseen scenarios and the inefficient modeling of robot-target relationships within the training data, resulting in a significant waste of training resources. To address these limitations, we present the \textbf{R}ecurrent \textbf{G}eometric-prior \textbf{M}ultimodal \textbf{P}olicy (\textbf{RGMP}), an end-to-end framework that unifies geometric-semantic skill reasoning with data-efficient visuomotor control. For perception capabilities, we propose the Geometric-prior Skill Selector, which infuses geometric inductive biases into a vision language model, producing adaptive skill sequences for unseen scenes with minimal spatial common sense tuning. To achieve data-efficient robotic motion synthesis, we introduce the Adaptive Recursive Gaussian  Network, which parameterizes robot-object interactions as a compact hierarchy of Gaussian processes that recursively encode multi-scale spatial relationships, yielding dexterous, data-efficient motion synthesis even from sparse demonstrations. Evaluated on both our humanoid robot and desktop robot, the RGMP framework achieves 87\% task success in generalization tests and exhibits 5× greater data efficiency than the state-of-the-art model. This performance underscores its superior cross-domain generalization, paving the way for more versatile and data-efficient robotic systems. 
\end{abstract}

\begin{links}
% \vspace{-0.2cm}
\link{Code}{https://github.com/xtli12/RGMP.git}
\end{links}

\begin{figure}[t]
  \centering
    \includegraphics[width=\linewidth]{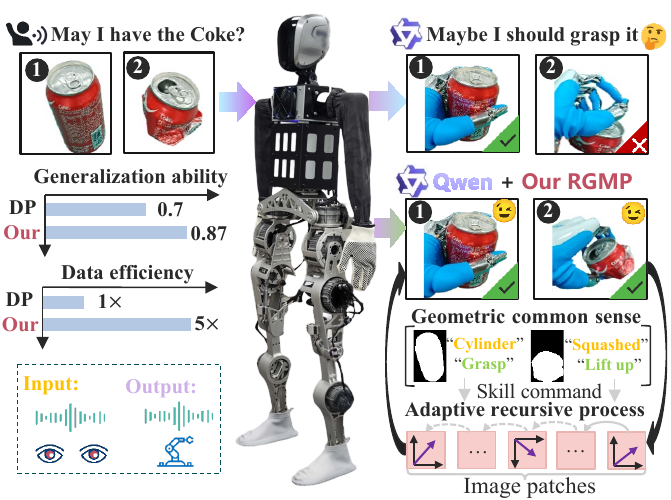}
  % \vspace{-0.7cm}
   \caption{\textbf{Overview of our framework.} By applying semantic cues from human instructions with common sense information derived from visual perception, our RGMP formulates the robot-targets spatial relationships for tasks. RGMP achieves an 8\% performance improvement and exhibits 5× greater data efficiency than Diffusion Policy.}
     % \vspace{-0.6cm}
   \label{fig:humanoid}
\end{figure}

\section{Introduction}
Humanoid robots demonstrate substantial potential in performing diverse human-level tasks, ranging from adaptive decision-making to complex manipulation~\cite{tong2024advancements,li2024okami}. However, current research predominantly relies on data-driven approaches, which require extensive training datasets to achieve robust multimodal decision-making and generalizable visuomotor control~~\cite{zitkovich2023rt,liu2024rdt,intelligence2025pi05visionlanguageactionmodelopenworld}. While these methods show promise in task-specific applications, they often overlook geometric reasoning and spatial awareness, limiting the ability of robots to perceive contextually under unseen environments. 
Thus, developing data-efficient and reasoning-capable methods is imperative to achieve the context-aware, adaptable humanoid systems required for real-world applications~\cite{abu2025development,skubis2023humanoid}.
% Consequently, there is a pressing need for approaches that combine data-efficient techniques with reasoning capabilities to enable more context-aware, and adaptable humanoid systems for real-world applications~\cite{abu2025development,skubis2023humanoid}.

Traditional Vision-Language Models (VLMs) such as PaLM-E~\cite{driess2023palm} and InstructBLIP~\cite{liu2024improved} demonstrate remarkable capabilities in parsing semantic intent from language-vision inputs. These models leverage large-scale pretraining to generate task plans conditioned on visual observations, yet their ability to associate abstract instructions with contextually appropriate robotic skills remains constrained~\cite{team2024octo}. For instance, those models struggle to resolve ambiguities in skill selection (e.g., grasping vs. pinching) when confronted with targets of varying shapes in unseen scenes. This limitation stems from insufficient integration of spatial object geometry (e.g., bounding boxs, shapes) with semantic task specifications, which is a gap exacerbated in dynamic environments where skill feasibility depends on generalized spatial reasoning~\cite{rothert2024sim}. Given this context, a fundamental question is: \textbf{I)} \textit{How can robots leverage spatial-geometric reasoning to enable feasible skill selection?}

Meanwhile, learning precise action policies from limited demonstrations remains an open challenge. While diffusion models~\cite{chi2023diffusion} and transformer-based architectures~\cite{c:22} have shown promise in trajectory generation, their reliance on extensive training data (10k+ trajectories) and computational complexity (1–5 Hz inference rates~\cite{zitkovich2023rt}) limits practical deployment. Imitation learning methods~\cite{zhang2018deep} partially mitigate this by leveraging human priors, but they often overfit to demonstration-specific features, achieving merely 40–60\(\%\) success rates on unseen objects~\cite{liang2023code}. The crux lies in disentangling task-invariant visual features (e.g., context-based features) from task-specific motion patterns. Therefore, an additional intriguing question is: \textbf{II)} \textit{How can the inherent mechanisms of robot learn the generalized ability with limited demonstrations?} 

To bridge these critical gaps, we introduce the \textbf{RGMP} (\textbf{R}ecurrent \textbf{G}eometric-prior \textbf{M}ultimodal \textbf{P}olicy), an end-to-end architecture that synergizes multimodal spatial-geometric reasoning with data-efficient visuomotor control. Regarding the issue of spatial-geometric reasoning discussed in \textbf{I)}, we present Geometric-prior Skill Selector (GSS): \textbf{the first framework to explicitly bridge geometric reasoning with semantic task planning} through a novel geometric-object decomposition mechanism. By incorporating geometric inductive biases into a VLM with minimal common-sense tuning, the GSS introduces a human-like decision-making process that mirrors how humans combine visual geometry and task semantics to select appropriate skills. Our geometric priors are plug-and-play, modular, and minimal (e.g., basic shape/affordance heuristics), requiring only 20 rule-based constraints for robust performance. On a humanoid robot, the GSS enables the manipulation of diverse objects in unseen scenes via geometric consistency checks, proving its effectiveness in real-world deployment.

Regarding the challenge of data efficiency discussed in \textbf{II)}, we propose Adaptive Recursive Gaussian Network (ARGN): a framework that dynamically models \textbf{spatial dependencies between the robot and targets by adaptively reconstructing spatial memory}. In robotics, the high cost and labor intensity of data collection often result in limited dataset sizes, which can lead to overfitting if the visual processing network lacks careful architectural design to uncover latent data relationships. To this end, our ARGN employs Rotary Position Embedding (RoPE) to establish an implicit association between each observed image patch and the final executed action. We then introduce recursive computation in the Spatial Mixing Block to progressively model global spatial relationships from the first to the last visual patch. This recursive global connection forms the \textbf{spatial memory} of observed images for the robot, enabling it to identify end-effector positions most relevant to task execution. However, recursive computation is prone to vanishing gradients, which increases training difficulty and requires substantial data to mitigate this issue. To address this, we propose an Adaptive Decay Mechanism that dynamically controls the decay rate of historical memory, preventing the loss of key spatial memories and adaptively amplifying the weights of task-critical patches. Furthermore, we utilize Gaussian Mixture Models (GMM) to fit six Gaussian distributions, approximating a series of motions controlled by distinct joints of a six-degree-of-freedom robotic arm. Our contributions are as follows: 
\begin{itemize} [leftmargin=*]
\item[\ding{182}] \textbf{\textit{A geometric-prior skill selector.}} We propose the GSS, which enhances a VLM with low-rank geometric adapters to select parameterized skills from a pretrained library. By infusing shape-level commonsense, GSS prioritizes skills that satisfy latent geometric constraints, enabling human-aligned reasoning without task-specific fine-tuning.
\item[\ding{183}] \textbf{\textit{A plug-and-play data-efficient visuomotor.}} We propose ARGN to  modulate latent representations via adaptive decay mechanisms and rotary embedding to capture directional spatial dependencies in a temporally-consistent latent space. A hierarchical fusion block retains multi-scale visual cues and feeds them into a Gaussian Mixture encoder that factorizes 6-DoF trajectories into a compact mixture of full-covariance, enabling explicit goal-conditional density modeling under severe data scarcity. 
\item[\ding{184}] \textbf{\textit{Comprehensive real-robot evaluation.}} Our RGMP undergoes rigorous evaluation on two physical robotic platforms, exhibiting robust performance by jointly coupling geometric-semantic reasoning with recursive Gaussian feature re-weighting. Compared to Diffusion Policy, RGMP achieves 87\% success rate in generalization tests and exhibits 5× greater
data efficiency.
\end{itemize}

 \begin{figure*}[t]
  \includegraphics[width=\textwidth]{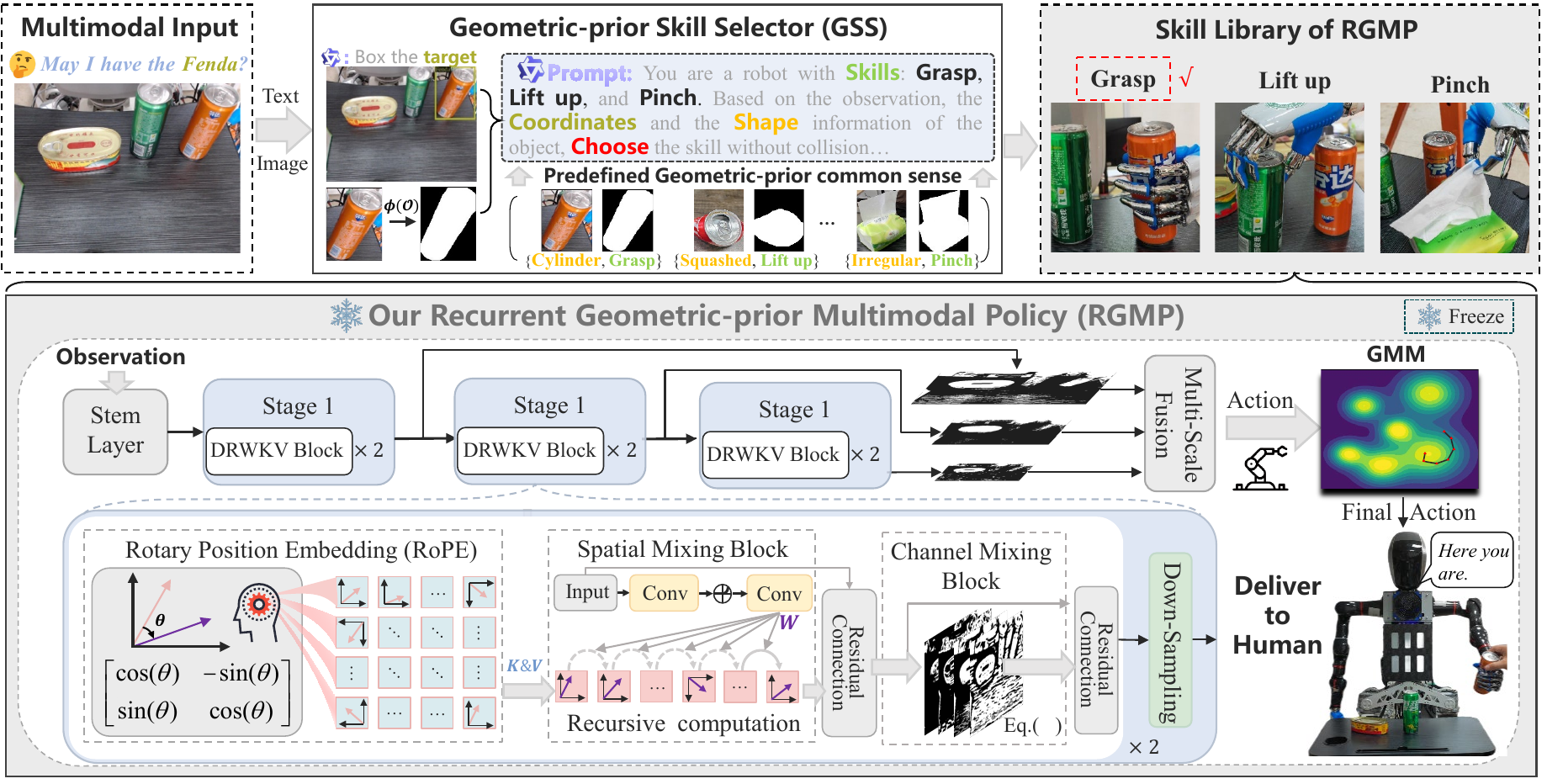}
  % \put(-423,59){\makebox(0,0)[l]{\fontsize{6}{6}\selectfont\ref{eq:rope}}}
  % \put(-245.2,52){\makebox(0,0)[l]{\fontsize{6}{6}\selectfont\ref{eq:adm}}}
  % \put(-248.3,18.3){\makebox(0,0)[l]{\fontsize{6}{7}\selectfont\ref{eq:wkv}}}
  \put(-166,18.5){\makebox(0,0)[l]{\fontsize{8}{8}\selectfont\ref{eq:channel}}}
  % \vspace{-0.3cm}
  \caption{\textbf{Pipeline of RGMP}. Upon receiving a speech command, the robot utilizes GSS to identify and localize the target object. By integrating object coordinates, shape cues (from Yolov8n-seg~\cite{yaseen2024yolov9} model \(\phi()\)), and geometric-prior knowledge, the robot selects an appropriate skill from the skill library, each associated with a pretrained RGMP model. The selected RGMP model then executes the task precisely through adaptive recursive feature extraction and GMM-based refinement. }
\label{fig:pipeline}
% \vspace{-0.6cm}
\end{figure*}

\section{Related Work}
\label{sec:related}

\subsection{Vision-Language Models}
% Refer to ``LLARVA:Vision-Action Instruction Tuning Enhances Robot Learning'' “Embodied AI with Two Arms: Zero-shot Learning, Safety and Modularity”. Introduce Q-wen vl.
Large Language Models (LLMs) are pivotal for robotic task planning, leveraging their capabilities in complex reasoning and response generation to autonomously create control code with loops, conditionals, and subroutines~\cite{ahn2022can}. This makes them well-suited for intricate perception-control tasks. Models like Palm-E~\cite{driess2023palm} exemplify this by integrating visual, linguistic, and robot state data to enable dynamic task execution~\cite{liang2023code}.
 % Large Language Models (LLMs) have become pivotal in robotic task planning, offering capabilities such as common sense reasoning, and language comprehension,  response generation~\cite{ahn2022can}. Palm-E~\cite{Driess2023PaLMEAE}, for instance, integrates visual, linguistic, and robot state data to facilitate dynamic task execution. LLMs can autonomously generate control code incorporating loops, conditionals, and subroutines, making them well-suited for complex perception-control integration tasks~\cite{liang2023code}. 
 Despite these advancements, current models still struggle to meet the diverse demands of real-world robotics. Recent progress in Vision-Language Models (VLMs) has significantly advanced vision-and-language integration tasks. Models like InstructBLIP~\cite{instructblip}, InstructGPT~\cite{InstructGPT_ouyang2022training}, LLaVA~\cite{liu2024visual,liu2024improved}, and PALM~\cite{Chowdhery2022PaLMSL}  leverage instruction tuning to improve image-text integration, setting new state-of-the-art benchmarks. However, their use in robotics applications faces considerable challenges due to real-world variability, platform heterogeneity, and the necessity for reliable action control~\cite{driess2023palm,shridhar2023perceiver,team2024octo,huang2025keeping}. These challenges often result in suboptimal performance in highly dynamic environments. To this end, we introduce the GSS to fuse  the Qwen-VL~\cite{yang2024qwen2} with low-rank geometric adapters to dispatch parameterized manipulation skills from a pre-computed library using common-sense priors.  Compared to previous models, GSS effectively manages uncertainties and dynamic requirements, making it a stable solution for robotic tasks. 

\subsection{Learning-Based Action Generation}
% Refer to ``Scaling Up and Distilling Down: Language-Guided Robot Skill Acquisition''.
Sophisticated manipulation systems drive the shift toward learning-based motion planning~\cite{team2024octo,liu2024rdt,SPIDER_ICML25,NRCA_ICML25}. However, those methods encounter three primary constraints: systematic dependence on predefined motion primitives~\cite{cruciani2018dexterous,cruciani2019dualarm}, inadequate cross-domain adaptability~\cite{liang2021learning}, and inherent complexities in reward formulation~\cite{kim2023pre,zeng2018learning,xu2021efficient}. 
Imitation learning is effective in physical deployments~\cite{zhang2018deep,Haldar-RSS-23,Li2-RSS-23,bogdanovic2020learning}, though its performance is bound to demonstration fidelity and scalability issues~\cite{decisiontransformer,xu2022prompting}. Emerging diffusion-based generative frameworks have shown potential for robotic decision-making, utilizing multi-stage probabilistic optimization for trajectory synthesis~\cite{zhang2023adding,Yoneda-RSS-23,huang2023diffusion,li2024source}. 
However, slow inference from sequential reverse diffusion precludes time-sensitive applications~\cite{dong2024diffuserlite}. Our RGMP framework overcomes this by merging objective-aware action synthesis with statistical motion modeling. It avoids iterative denoising while maintaining robustness, enabling efficient multimodal inference. The hierarchical architecture further unifies task planning and motor control via distilled action primitives and adaptive distributions.
% However, their practical adoption faces constraints arising from suboptimal inference speeds, predominantly caused by the temporal latency of sequential reverse diffusion processes, rendering them unsuitable for time-sensitive applications~\cite{dong2024diffuserlite}. To overcome these challenges, we introduce RGMP that synergizes objective-aware action synthesis with statistical motion modeling. Unlike conventional diffusion architectures, our vision-guided planner implements computationally efficient multimodal action inference, eliminating resource-intensive iterative denoising while maintaining decision robustness. The proposed hierarchical architecture achieves seamless visuomotor integration through distilled action primitives and adaptive probability distributions, effectively unifying strategic task decomposition with dynamic actuator coordination.

\section{Methodology}

Our RGMP framework (as shown in Algorithm~\ref{alg:RGMP}) integrates two components: the GSS, which translates verbal commands and visual cues into executable skills using geometric commonsense, and the ARGN, which processes visual inputs to predict manipulation actions. The policy learns to infer 3D spatial relationships directly from RGB by associating visual cues with actions, relying on an efficient implicit representation instead of explicit 3D reconstruction. 
% The complete pipeline is outlined in Algorithm~\ref{alg:RGMP}.

% Our RGMP consists of two key components: the GSS and the ARGN. The GSS translates verbal commands and visual cues into executable skill sequences using geometric-prior common sense, while the ARGN leverages a pretrained skill model and processes the visual inputs to predict the actions required for robotic manipulation. We train the policy to decode implicit geometric information from RGB by associating 3D spatial cues with robotic action labels and commonsense reasoning, avoiding costly explicit 3D reconstruction in favor of an efficient implicit representation. The whole pseudocode for our pipeline is presented in Algorithm~\ref{alg:RGMP}.

\subsection{Geometric-prior Skill Selector}

\textbf{Motivation.} A key challenge in robotics is fine-grained skill selection (e.g., grasping vs. pinching) for diverse-shaped targets or in unseen scenes. Traditional VLMs, despite enabling object recognition and localization, fail to map semantic observations to accurate actions due to overlooking \textbf{geometric priors} in vision-action mapping. This motivates our pioneering GSS framework, which bridges geometric reasoning and semantic task planning via a novel geometric-object decomposition mechanism.

The GSS comprises two stages. In the first stage, a VLM~\cite{bai2023qwen} is utilized to interpret human commands, enabling the robot to identify and localize the target object within the observed image. In the second stage, based on the bounding box obtained from the first stage, the system analyzes the target object’s common sense information, including its relative position and its shape information. Subsequently, the system selects the pretrained skill model from a skill library according to the output of the GSS. The planning function operates through:
\begin{equation}\small
\label{eq:vlm}
\mathcal{P} = plan(\mathcal{I},{\kern 1pt} {\kern 1pt} \mathcal{O}{\kern 2pt}|{\kern 2pt}\mathcal{C}\;),
\end{equation}
% \vspace{-0.2cm}
where \(\mathcal{P}\) is the generated action plan, \(\mathcal{I}\) denotes the current user instruction, \(\mathcal{O}\) is a current visual observation, and \(\mathcal{C}\) represents a predefined context (instruction, prompt, and common sense) that consists of n examples $\{({\mathcal{I}_i},{\mathcal{O}_i},{\mathcal{P}_i})\}_{i = 1}^n$, enabling in-context learning.  

Specifically, the observation  \(\mathcal{O}\) is an RGB image annotated with a bounding box by the VLM. Subsequently, the VLM generates an executable skill based on the predefined context \(C\) and the geometric-prior common sense, which includes relative position and the shape information. For example, when the instruction is \textit{\enquote{I want Fanta}}, our pipeline adheres to the context \textit{\enquote{Please box the target object in the instruction}} to identify the \textit{\enquote{Fanta}} can among various other items and apply Yolov8n-seg to obtain the shape information of Fanta.  The VLM subsequently synthesizes operational directives by integrating its established contextual framework \(C\) with geometric-based prior reasoning. Our GSS is plug-and-play, modular, and minimal (for implementation details, please refer to Appendix A in the code repository).

\begin{algorithm}[!t]
\caption{The RGMP Framework}
\label{alg:RGMP}
% \SetAlgoLined
\SetNoFillComment
\SetArgSty{textnormal}
\small{\KwIn{Training epochs $E$, conversation round $T$, human speech $\mathcal{I}$, human demonstrations $\mathcal{D}$ with capacity $M$, VLM model $Q$, RGMP $\mathcal{G}_{m}$ }}
\small{\KwOut{Actions of robot  $a^{*}$}}

 \BlankLine
 % {\color{DarkBlue}{\tcc{Human Demonstration Collecting:}}}
\For {$i=1, 2, ..., M$ }{
    $d_{i}\leftarrow(\mathcal{O}_{i}, \mathcal{J}_{i})$ through Eq.~({\ref{eq:collection}})  
    
    $\mathcal{D} \leftarrow d_{i}$}
    
    return $\mathcal{D}$

\BlankLine
{\tcc{RGMP Training pipeline:}}
\For {$e=1, 2, ..., E$ }{
    \BlankLine
    % {\color{DarkBlue}{\tcc{Updating ARGN parameter}}}
    
    $F_{0}\leftarrow Stem(\mathcal{O}_{i})$
    % by Eq.~({\ref{eq:stem}}) 

    ${\mathcal{W}},{K_s}, {V_s}  \leftarrow \mathcal{A}(F_{0})$, $\mathcal{R}(F_{0})$ by Eq.~({\ref{eq:adm}}))

    ${F_{1},F_{2},F_{3}}  \leftarrow \mathcal{S}({K_s}, {V_s}, {\mathcal{W}})_{\times 3}$

    ${a_{in}}  \leftarrow \mathcal{M}(F_{1},F_{2},F_{3})$ by Eq.~({\ref{eq:ain}})

    $\mathcal{L} \leftarrow  ({a_{in}},{a_{ground}})$ through Eq.~({\ref{eq:loss}}) 

    $\mathcal{G}_{m}  \leftarrow \mathcal{G}_{m}  - \eta \nabla \mathcal{L}$
    \BlankLine
    % {\color{DarkBlue}{\tcc{Updating GMM parameter}}}
    $\Theta \leftarrow \mathcal{J}_{i} $ in $\mathcal{D}$ through Eq.~({\ref{eq:GMM}})    
    }
    return $\mathcal{G}_{m}$, $\Theta$
 \BlankLine

{\tcc{Inferencing pipeline:}}

\For {$t=1, 2, ..., T$ }{
    \BlankLine
    % {\color{DarkBlue}{\tcc{Human play the chime and speak}}}
    % $\mathcal{I}, box(x_1,y_1,x_2,y_2)\leftarrow$ human speech, Qwen($\mathcal{O}$) 
    % \BlankLine
    % {\color{DarkBlue}{\tcc{VLM generate the response plan}}}
    $ Box(x_1,y_1,x_2,y_2) \leftarrow \mathcal{Q}(\mathcal{I}, \mathcal{O}| \mathcal{C})$ by Eq.~({\ref{eq:vlm}})

    $\mathcal{O}_{s} \leftarrow  (x_1,y_1,x_2,y_2),\kern 2pt \phi_{a}(\mathcal{O},Box)$

    $\mathcal{P} \leftarrow \mathcal{Q}(\mathcal{I}, \mathcal{O}_{s}| \mathcal{C})$
    
    Voice $\leftarrow$ response in $\mathcal{P}$
    
    \BlankLine
    % {\color{DarkBlue}{\tcc{RGMP generate the action plan}}}        
    \If {'Skill' in $\mathcal{P}$}{
    $a^{*}\leftarrow \mathcal{G}_{m}(\mathcal{O})$ through Eq.~({\ref{eq:regression}}) 
    
    }

    return $a^{*}$  
    }
% \vspace{-0.1cm}
\end{algorithm}

\subsection{Adaptive Recursive Gaussian Network}
\label{subsec:drwkv}
\textbf{Motivation.}  In robotic tasks, understanding spatial relationships from the robot's visual perspective is essential. The robot must identify which parts of the scene correspond to the position of its end-effector. Previous methods often struggle to uncover the underlying relationships between different image regions in unseen scenes due to inherent limitations in visuomotor representation learning, which limits generalization capability. To address this issue, we propose the ARGN framework, which is designed to adaptively model comprehensive \textbf{spatial dependencies} between the robot and target objects in unseen environments, while mitigating overfitting in scenarios with limited training data.

\begin{figure}[t]
% \Centering
  \includegraphics[width=\columnwidth]{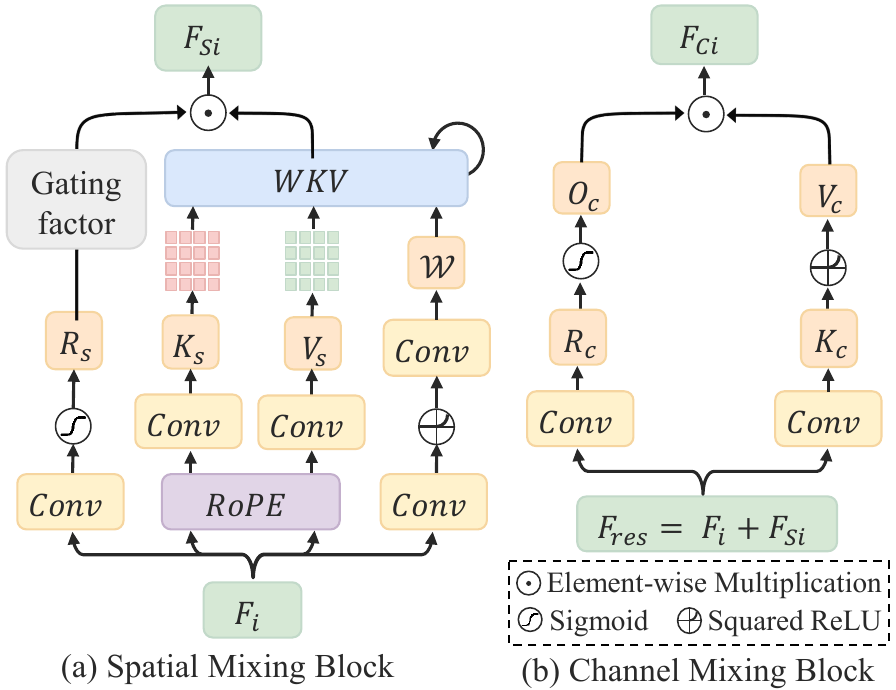}
  % \vspace{-0.6cm}
  % \captionsetup{justification=justified, singlelinecheck=false}
  \caption{\textbf{Structure of (a) Spatial Mixing Block and (b) Channel Mixing Block.} 
  The Spatial Mixing Block integrates an ADM for Dynamic Decay $\mathcal{W}$ and RoPE for directional awareness, enhancing spatial aggregation. The Channel Mixing Block reallocates channel-wise feature responses by integrating correlations between channels.  }
  
  % (a) The Spatial Mixing Block integrates an ADM module to get Dynamic Decay $\mathcal{W}$ during recursive operations, and employs RoPE to introduce directional awareness relative to spatial positions. These mechanisms collectively enhance the ability of block to integrate spatial information effectively. (b) The Channel Mixing Block reallocates channel-wise feature responses by integrating correlations between channels.

\label{fig:RWKV}
% \vspace{-0.4cm}
\end{figure}

In our framework, we apply recursive operation to establish global connection, which establishes the\textbf{ spatial memory} of observed images. This memory mechanism enables the identification of end-effector positions most relevant to task execution. However, recursive computation inherently suffers from vanishing gradients, increasing training difficulty and demanding substantial data to mitigate this limitation. To address this, we propose an Adaptive Decay Mechanism (ADM) to dynamically \textbf{control historical memory decay rates} to prevent the vanishing of key spatial memories, and to adaptively amplify weights for task-critical patches. In Stage 1, the input \( F_0 \) is processed by the Spatial Mixing Block, where the ADM generates content-adaptive decay factors \( \mathcal{W} \) to regulate memory retention.
\begin{equation}\small
\label{eq:adm}
\mathcal{W} = \sigma(\mathcal{C}_{1\times1}(SReLU(\mathcal{C}_{3\times3}(F_0)))),  
\end{equation}
where \(\mathcal{C}_{1\times1}(\cdot)\) represents a convolutional operation with \(1\times1\) kernel that enables channel re-calibrate, \(\sigma(\cdot)\) denotes the Sigmoid activation function. RoPE is then applied to encode positional information through rotational transformations, enhancing sensitivity to relative spatial offsets without learnable position parameters. After applying RoPE, we slice \(K_s\) and \(V_s\) into  \(16 \times 16\) image patches (as illustrated in Fig.~\ref{fig:RWKV}), and then employ recursive computation in the Spatial Mixing Block to progressively model global spatial relationships from the first visual patch to the last:
\begin{equation}\small
\label{eq:wkv}
\begin{gathered}
        WKV_{i} = \frac{n_{i} + e^{u} \odot k_{i} \odot v_{i}}{d_{i} + e^{u} \odot k_{i}}, \\
        \underbrace{{n_{i} = n_{i-1} \odot e^{-\mathcal{W}} + k_{i} \odot v_{i}}}_{\textbf{Cumulative memory of } k_{i} \odot v_{i}}, 
        \underbrace{{d_{i} = d_{i-1} \odot e^{-\mathcal{W}} + k_{i}}}_{\textbf{Cumulative memory of } k_{i}},
    \end{gathered}
\end{equation}
where  \( i \in  [0,  (H\times W)/(16\times 16) ) \), \(k_{i}\) and \(v_{i}\) represent the patches of \(K_s\) and \(V_s\), respectively. The initial values \(n_{0}\) and \(d_{0}\) are copied from \(k_{0}\). The parameter \(u\in  (0,1)\) denotes the learnable position compensation, which enhances the sensitivity of the model to local positions. The term \(\mathcal{W}\) represents content-adaptive decay factors that control the decay rate of historical memory (as shown in Equation~\ref{eq:adm}). Finally, a dynamic weight is generated through the gating factor \(R_s\) to modulate the contribution of the output from the Spatial Mixing Block to the current state:
\begin{equation}\small
F_{S0} = \sigma(\mathcal{C}_{1\times1}(F_0))\odot WKV.
\end{equation}
Then, \(F_{S0}\) is residually connected with \(F_0\) to obtain \(F_{res}\) (as shown in~Fig.\ref{fig:RWKV}). We apply the Channel Mixing Block to reallocate channel-wise feature weights for feature extraction:
\begin{equation}\small
\label{eq:channel}
F_{C0} = \sigma(\mathcal{C}_{1\times1}(F_{res}))\odot (SReLU(\mathcal{C}_{3\times3}(F_{res})))
\end{equation}
\(F_1\) is obtained after down-sampling the output of two ARGN blocks using a \(3 \times3\) convolutional operation. Subsequent stages (Stage 2–3) repeat this process, and multi-scale features \( F_1, F_2, F_3 \) are fused via learnable weights:
\begin{equation}\small
F_f =\alpha_1(\mathcal{C}_{1\times1}(F_{1}))+\alpha_2(\mathcal{C}_{1\times1}(Up(F_{2})))+\alpha_3(Up(F_{3})),
\end{equation}
where \(F_i\) denotes the feature map processed by the \(Stage_i\) (\(i=1, 2, 3\)), and \(\alpha_i\) are the learnable parameters that assign weights to feature maps of different levels during the feature fusion process. We then generate the initial predicted action \(a_{in}\) based on the fused feature map \(F_f\) as follows:
\begin{equation} \small
\label{eq:ain}
    {a_{in}} = Linear(\mathcal{C}_{3\times3}(F_f))),
\end{equation}
To minimize the mean-squared error (MSE) between the predicted action and the ground truth action, we use the following loss function:
\begin{equation}
\label{eq:loss}
    \mathcal{L} = {\mathop{\rm MSE}\nolimits} ({a_{in}},{a_{ground}}),
\end{equation}
where \(\mathcal{L}\) represents a loss function, \({a_{ground}}\) is the ground truth action from human demonstrations (for detailed formulations of ARGN, please refer to Appendix B).
\paragraph{Why Gaussian Mixture Model?} 
When using a single Gaussian~\cite{chi2023diffusion}, the model tends to regress to the mean, suppressing distinct action modes and leading to suboptimal control accuracy. In contrast, a Gaussian Mixture Model (GMM) enables the modeling of separate action clusters, each with its own mean and covariance, allowing for more accurate representation of the action distribution.
 Let \( \mathbf{x} \in \mathbb{R}^n \) denote ground-truth joint configurations. The GMM uses \( K=6 \) components with prior \( \alpha_k \), mean \( \mu_k \), and covariance \( \Sigma_k \), with probability density:
\begin{equation}
\label{eq:GMM}
P(\mathbf{x} \mid \Theta) = \sum_{k=1}^{K} \alpha_{k} \mathcal{N}\left(\mathbf{x} \mid \mu_{k}, \Sigma_{k}\right),
\end{equation}
where \( \mathcal{N} \) is a multivariate Gaussian. Parameters \( \{\alpha_k, \mu_k, \Sigma_k\} \) are estimated via the EM algorithm to maximize data likelihood, capturing latent joint space structures. The initial prediction \( a_{\text{in}} \) is compared to GMM clusters using Mahalanobis distance:
\begin{equation}\small
l_{k} = \sqrt{\left(a_{\text{in}} - \mu_{k}^{\omega}\right)^{T}\left(\Sigma_{k}^{\omega, \omega}\right)^{-1}\left(a_{\text{in}} - \mu_{k}^{\omega}\right)},
\end{equation}
where \( l_k \) measures distance to the \( k \)-th component. The final action \( a^{*} \) is the closest cluster center:
\begin{equation}\small
\label{eq:regression}
a^{*} = \arg \min_{\mu_{k}^{\omega}} l_{k},
\end{equation}
where \(a^{*}\) is the final predicted action (For detailed derivations, please refer to Appendix B and C).

\begin{figure*}[t]
  \includegraphics[width=\textwidth]{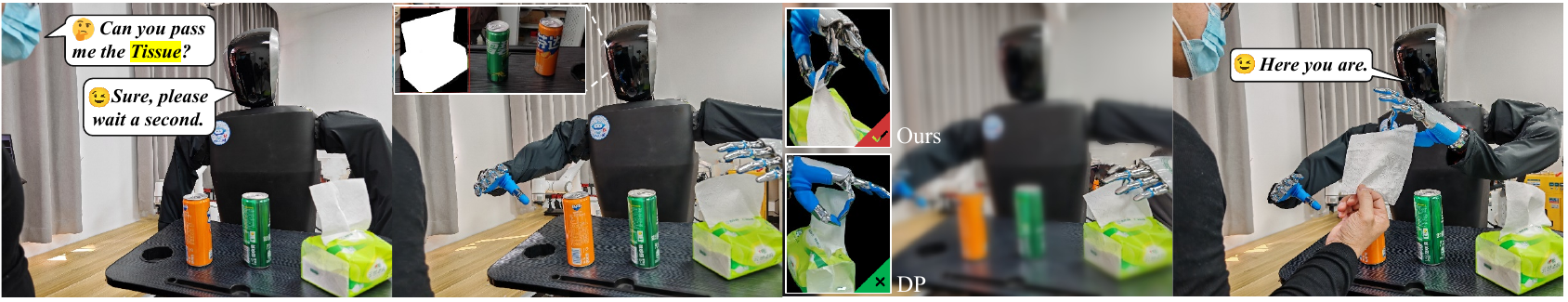}
  % \vspace{-0.6cm}
  \caption{\textbf{Pipeline of human-robot interactions}. We validate models on the task of ``passing me the tissue'', with a training dataset comprising only 40 instances of tissue pinching actions. Our RGMP performs better than DP (Diffusion Policy).}
\label{fig:experiment1}
% \vspace{-0.2cm}
\end{figure*}

\begin{table*}[!t]
    \centering
    % \vspace{-2pt}
    \label{tab:results}
    \resizebox{0.95\linewidth}{!}{
    \begin{tabular}{c||ccc|ccc|ccc|ccc|ccc}
\hline\thickhline
\rowcolor{mygray}
        & \multicolumn{3}{c}{Fanta} & \multicolumn{3}{c}{Sprite} & \multicolumn{3}{c}{Tissue} & \multicolumn{3}{c}{Squashed Coke} & \multicolumn{3}{c}{Human Hand} \\
        \rowcolor{mygray}
      \multirow{-2}{*}{Methods}  & $Acc_s$ & $Acc_t$ & $Acc \uparrow$ & $Acc_s$ & $Acc_t$ & $Acc \uparrow$ & $Acc_s$ & $Acc_t$ & $Acc \uparrow$ & $Acc_s$ & $Acc_t$ & $Acc \uparrow$ & $Acc_s$ & $Acc_t$ & $Acc \uparrow$ \\
\hline\hline
\multicolumn{16}{l}{with \textit{ResNet50}} \\
\hline
Qwen-VL & 0.65 & 0.54 & 0.35 & 0.60 & 0.42 & 0.25 & 0.65 & 0.46 & 0.30 & 0.65 & 0.46 & 0.30 & 0.70 & 0.57 & 0.40 \\
GSS & 0.85 & 0.53 & 0.45 & 0.75 & 0.46 & 0.35 & 0.85 & 0.47 & 0.40 & 0.85 & 0.47 & 0.40 & 0.85 & 0.56 & 0.48 \\
\hline\hline
\multicolumn{16}{l}{with \textit{Transformer}} \\
\hline
Qwen-VL & 0.60 & 0.58 & 0.35 & 0.65 & 0.54 & 0.30 & 0.70 & 0.50 & 0.35 & 0.60 & 0.58 & 0.35 & 0.65 & 0.62 & 0.40 \\
GSS & 0.80 & 0.56 & 0.45 & 0.75 & 0.53 & 0.40 & 0.85 & 0.53 & 0.45 & 0.85 & 0.53 & 0.45 & 0.85 & 0.64 & 0.54 \\
\hline\hline
\multicolumn{16}{l}{with \textit{ManiSkill2-1st}} \\
\hline
Qwen-VL & 0.70 & 0.57 & 0.40 & 0.65 & 0.69 & 0.45 & 0.65 & 0.53 & 0.34 & 0.65 & 0.54 & 0.35 & 0.65 & 0.62 & 0.40 \\  
GSS & 0.85 & 0.53 & 0.45 & 0.80 & 0.68 & 0.54 & 0.85 & 0.53 & 0.45 & 0.80 & 0.56 & 0.45 & 0.85 & 0.70 & 0.60 \\
\hline\hline
\multicolumn{16}{l}{with \textit{Diffusion Policy}} \\
\hline
Qwen-VL & 0.65 & 0.76 & 0.49 & 0.65 & 0.75 & 0.50 & 0.65 & 0.68 & 0.44 & 0.65 & 0.62 & 0.40 & 0.70 & 0.71 & 0.50 \\ 
GSS & \textbf{0.85} & \textbf{0.76} & \textbf{0.65} & \textbf{0.80} & \textbf{0.77} & \textbf{0.62} & \textbf{0.85} & \textbf{0.69} & \textbf{0.59} & \textbf{0.85} & \textbf{0.65} & \textbf{0.55} & \textbf{0.90} & \textbf{0.83} & \textbf{0.74} \\
\hline\thickhline
    \end{tabular}
    }
    % \vspace{-0.2cm}
    \caption{\textbf{Ablation study of GSS and Qwen-VL}. Experiments use the scenes with Fanta, Sprite, and tissue paper (objects repositioned randomly per trial). Flattened Coke cans and human hands were tested separately. Each skill category included 40 training demonstrations, with test results from 20 random repositioning trials.}
    \label{tb:real}
    % \vspace{-0.4cm}
\end{table*}

\begin{table}[t]
% \vspace{0.1cm}
\footnotesize
\centering
{

\resizebox{1\columnwidth}{!}{
\setlength\tabcolsep{3pt}
\renewcommand\arraystretch{1.05}
\begin{tabular}{c||cccc|c}

\hline\thickhline
\rowcolor{mygray}
Methods& Fanta$\uparrow$   &  Coke $\uparrow$ & Spray $\uparrow$ & Hand $\uparrow$ & Average $\uparrow$  \\ \hline\hline
{ManiSkill2-1st} &  0.70 & 0.60 &0.63 & 0.62 & 0.64\\ 
{Octo} &  0.65 & 0.55 &0.58 & 0.62 & 0.60 \\
{OpenVLA} &  0.68 & 0.58 &0.61 & 0.60 & 0.62\\
{RDT-1b} &  0.70 & 0.61 &0.60 & 0.62 & 0.64\\
{Diffusion Policy} &  0.75 & 0.65 &0.68 & 0.72 & 0.70 \\
{Dex-VLA} &  0.87 & 0.66 &0.71 & 0.84 & 0.77 \\
\hline
\textbf{RGMP(ours)}  &  \textbf{0.98} & \textbf{0.78} &\textbf{0.81} & \textbf{0.90} & \textbf{0.87}
% \vspace{-0.6cm}

\end{tabular}
}}
% \vspace{-8pt}
\caption{\textbf{Evaluation results of generalized manipulation capability.} Models are only trained on 40 Fanta can grasping demonstrations. Metrics indicate grasping success rates for Fanta cans, Coke cans, spray bottles, and human hands.}
% \vspace{-0.4cm}
\label{tb:grasp}
\end{table}

\section{Experiments}
We evaluate the effectiveness and generalization of the RGMP framework by assessing its core components (GSS and ARGN) and benchmarking it against state-of-the-art methods. The evaluation metrics, experimental setup, implementation details, and comprehensive results are as follows:
% In this section, we evaluate the effectiveness and generalized ability of the proposed RGMP framework. Our experiments are designed to assess the performance of each core component, including the GSS and ARGN. We also compare our approach against state-of-the-art baselines in various manipulation tasks. The evaluation metrics, experimental setup, implementation details, and comprehensive results are detailed as follows.
\subsection{Hardware Setup}
\label{exp:setup}
We conduct experiments on two robotic platforms: a humanoid robot, with evaluations focused on the upper limb (please see Appendix D for details), and a desktop dual-arm robot, designed to test cross-embodiment generalization capability. The desktop robot is equipped with an RGB camera and two 6-DoF arms for manipulation tasks.
\subsection{Dataset and Evaluation Criteria}
\label{subsec:data}
To validate the effectiveness of the RGMP, we collected 120 trajectories for the skill library. Each trajectory corresponds to an execution path associated with an RGB image captured prior to the robotic arm performing an action: 
\begin{equation}\small
\label{eq:collection}
d_{i}=(\mathcal{J,O}), 
\end{equation}
where $\mathcal{J}$ denotes the joint space of the robotic arm, each trajectory specifies the motion of the arm from its initial configuration to the target spatial location and end-effector pose. In real-world evaluations, the model performance is assessed using two complementary metrics. The skill success rate, denoted as $Acc_s$, is recorded when the robot correctly identifies and selects the appropriate skill for the task. Additionally, the execution accuracy $Acc_t$ quantifies the precision with which the robot executes the selected skill to retrieve the target object. Consequently, the final success rate $Acc$ is defined as the product of these two metrics:
\begin{equation}\small
\label{eq:acc}
Acc=Acc_s \times Acc_t, 
\end{equation}
The detailed criteria for ManiSkill2 manipulation tasks can be referred to in Appendix E in the code link.
% ~\ref{ap:mani} 
% in supplements.

\begin{figure*}[t!]
  \includegraphics[width=\textwidth]{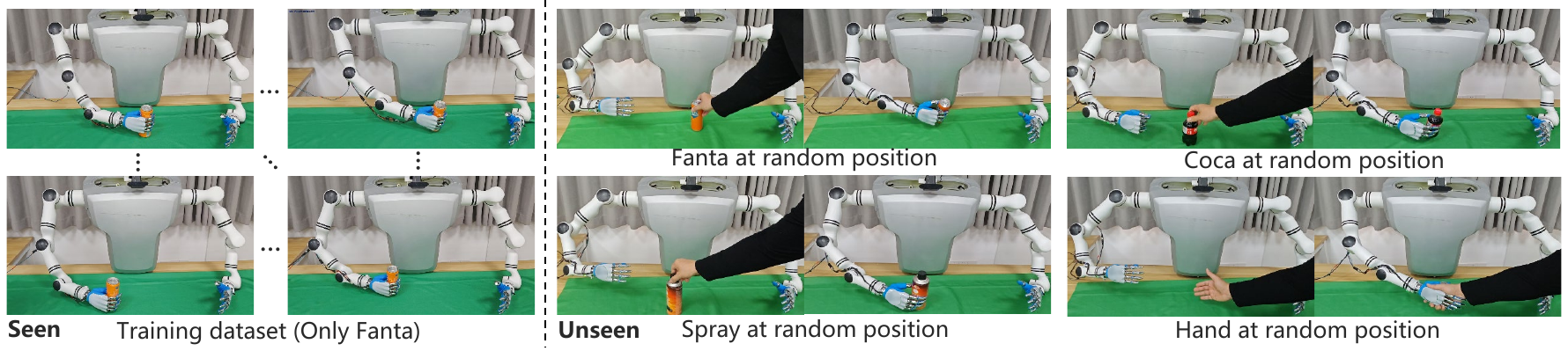}
  % \vspace{-0.7cm}
  \caption{\textbf{Generalization ability of RGMP.} We test RGMP on grasping various unseen objects at random positions. Despite being trained on only 40 demonstrations of grasping a Fanta, RGMP reliably grasped the can from any position and generalized this proficiency to unseen objects like a Coke bottle, a spray can, and human hand, demonstrating remarkable versatility.
  % Notably, our RGMP was trained on a dataset comprising only 40 demonstrations of grasping a Fanta can. RGMP exhibited proficient performance in grasping the Fanta can from any position. Furthermore, it displayed significant generalization capabilities to other unseen objects, including a Coke, a Spray, and a Hand, underscoring its versatility and adaptability in manipulation. 
  }
\label{fig:experiment2}
% \vspace{-0.4cm}
\end{figure*}

\begin{figure}[t]
% \Centering
  \includegraphics[width=\columnwidth]{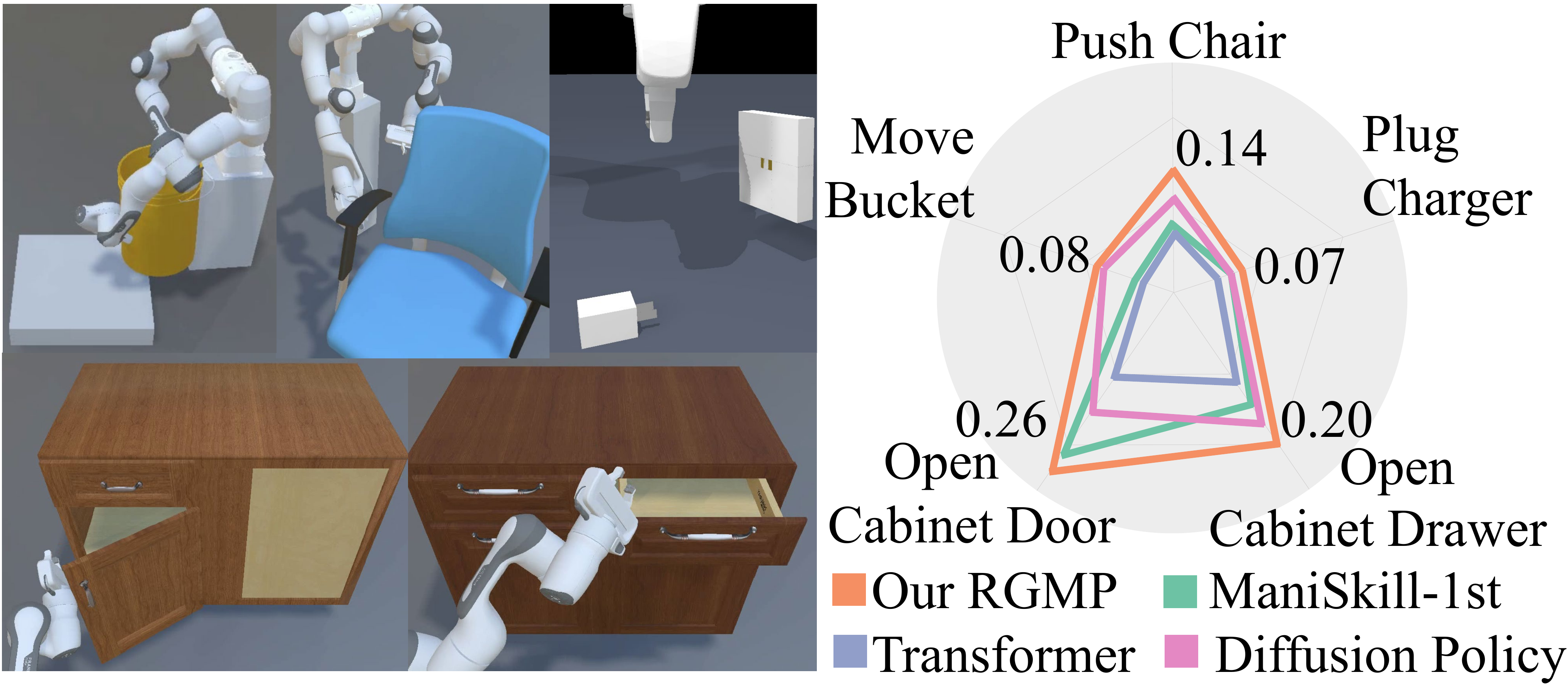}
  % \vspace{-0.7cm}
  % \captionsetup{justification=justified, singlelinecheck=false}
  \caption{\textbf{Performance on ManiSkill2 simulator.} We assess the effectiveness of RGMP and SOTA models across five manipulation tasks of ManiSkill2.}
\label{fig:mani}
% \vspace{-0.6cm}
\end{figure}

\subsection{Performance Comparison and Ablation Study}
\label{subsec:performance}
To evaluate RGMP, we conducted real-world comparative experiments against ResNet50~\cite{he2016deep}, Transformer~\cite{c:22}, the first-place entry~\cite{gao2023two} in the ManiSkill2 challenge, Octo~\cite{team2024octo}, OpenVLA~\cite{kim2024openvla}, RDT-1b~\cite{liu2024rdt}, Dex-VLA~\cite{wen2025dexvla}, and Diffusion Policy. Experiments involved random target object placement, with success defined as accurate instruction understanding, correct manipulation execution (object delivery to humans), and collision avoidance. To assess generalization and cross-domain transferability, we deployed the trained model on a desktop dual-arm platform, using a low-data setup: 40 interaction samples for Fanta can grasping as the exclusive training data, with evaluation on three unseen categories (human hands, spray bottles, Coke cans) at random workspace positions.
Tables~\ref{tb:real} and~\ref{tb:grasp} show RGMP outperforms baselines across tasks, with top \(Acc\), \(Acc_s\), and \(Acc_t\) for Fanta cans, Sprite cans, tissue papers, deformed Coke cans, and human hands, validating its effectiveness on regular/irregular objects. As Table~\ref{tb:real} demonstrates, our GSS yields a 15-25\% accuracy improvement in skill selection compared to Qwen-VL. Ablation studies (Table~\ref{tb:abla1}) confirm that integrating GMM with ARGN enhances performance: for Diffusion Policy, GSS+GMM yields a 0.55 $Acc$ versus 0.49 without GMM, while ARGN with GSS+GMM achieves a 0.69 $Acc$ in picking squashed Coke, demonstrating the effectiveness of GMM in refining predictions. Additionally, Table~\ref{tb:abla2} validates the contributions of RoPE, Spatial Mixing Blocks (SMB), and Channel Mixing Blocks (CMB): their combined use yields the highest accuracy across all objects (0.98 for Fanta, 0.78 for Coke, 0.81 for spray, 0.90 for hands). Beyond three primitives (grasp/lift-up/pinch), we evaluate five non-grasp ManiSkill2 tasks, complex tasks like plugging chargers (pinch) and opening cabinets (grasp) are dynamically composed from our atomic primitives.  As shown in Figure~\ref{fig:mani}, our RGMP achieves the highest performance across all tasks, demonstrating its transferability and generalization capability.
 Furthermore, as shown in Table~\ref{tb:effi}, RGMP achieves a score of 0.98 with 40 training samples, using 5$\times$ fewer samples than the 200 required by DP.

\section{Conclusion and Future Work}
\label{sec:conclution}
This work addresses semantic-spatial skill alignment and visuomotor overfitting in humanoid robotics via our RGMP, an end-to-end framework integrating GSS and ARGN. By dynamically associating contextual skills and decomposing 6-DoF trajectories into probabilistically regularized Gaussian components, RGMP achieves 87\% generalization success and 5$\times$ greater data efficiency than Diffusion Policy in human-robot interaction. Results show explicit neuro-symbolic coordination enables robust generalization across unseen objects/scenes, advancing collaboration with a scalable adaptive manipulation foundation.
Future work will explore functional generalization: demonstrating one primary object function allows automatic inference of trajectories for others, eliminating exhaustive teaching and enhancing efficiency in dynamic environments.

\begin{table}[!t]
\footnotesize
\centering
\resizebox{1\columnwidth}{!}{
\setlength\tabcolsep{3pt}
\renewcommand\arraystretch{1.05}
\begin{tabular}{c|c||ccc|ccc}
\hline\thickhline
\rowcolor{mygray}
 & & \multicolumn{3}{c|}{Tissue} & \multicolumn{3}{c}{Squashed Coke} \\
\rowcolor{mygray}
\multirow{-2}{*}{Methods} &\multirow{-2}{*}{GMM} & $Acc_s$ & $Acc_t$ & $Acc$ & $Acc_s$ & $Acc_t$ & $Acc$ \\
\hline\hline
 & -- & 0.85 & 0.58 & 0.50 & 0.80 & 0.61 & 0.49 \\
\multirow{-2}{*}{Diffusion policy} & $\checkmark$ & 0.80 & 0.68 & 0.56 & 0.85 & 0.65 & 0.55 \\
\hline
& -- & 0.80 & 0.69 & 0.55 & 0.85 & 0.71 & 0.60 \\
\multirow{-2}{*}{ARGN(ours)} & $\checkmark$ & \textbf{0.85} & \textbf{0.71} & \textbf{0.60} & \textbf{0.90} & \textbf{0.77} & \textbf{0.69} \\
% \hline\thickhline
\end{tabular}
}
% \vspace{-12pt}
\caption{\textbf{Ablation study of ARGN and GMM}. We validate models on the task of passing tissue and squashed Coke.}
% \vspace{-0.3cm}
\label{tb:abla1}
\end{table}

\begin{table}[!t]
\footnotesize
\centering
\resizebox{0.9\columnwidth}{!}{
\setlength\tabcolsep{3pt}
\renewcommand\arraystretch{1.05}
\begin{tabular}{ccc|cccc}
\hline\thickhline
\rowcolor{mygray}
RoPE & SMB & CMB & Fanta $\uparrow$ & Coke $\uparrow$ & Spray $\uparrow$ & Hand $\uparrow$ \\
\hline\hline
-- & $\checkmark$ & $\checkmark$ & 0.86 & 0.69 & 0.71 & 0.77 \\
$\checkmark$ & -- & $\checkmark$ & 0.83 & 0.75 & 0.76 & 0.82 \\
$\checkmark$ & $\checkmark$ & -- & 0.91 & 0.66 & 0.65 & 0.74 \\
$\checkmark$ & $\checkmark$ & $\checkmark$ & \textbf{0.98} & \textbf{0.78} & \textbf{0.81} & \textbf{0.90} \\
% \hline\thickhline
\end{tabular}
}
% \vspace{-10pt}
\caption{\textbf{Ablation study of the components of ARGN.} We evaluate RoPE, Spatial Mixing Block (SMB), and Channel Mixing Block (CMB) in grasping tasks.}
% \vspace{-0.3cm}
\label{tb:abla2}
\end{table}

\begin{table}[!t]
\footnotesize
\centering
\resizebox{0.8\columnwidth}{!}{
\setlength\tabcolsep{3pt}
\renewcommand\arraystretch{1.05}
\begin{tabular}{c||ccccc}
\hline\thickhline
\rowcolor{mygray}
Methods & 40 & 80 & 120 & 160 & 200 \\
\hline\hline
Diffusion Policy & 0.81 & 0.89 & 0.94 & 0.95 & 0.98 \\
\hline
\textbf{RGMP(ours)} & \textbf{0.98} & \textbf{0.98} & \textbf{0.99} & \textbf{0.99} & \textbf{0.99} \\
% \hline\thickhline
\end{tabular}
}
% \vspace{-10pt}
\caption{\textbf{Data efficiency comparison of RGMP and Diffusion Policy.} RGMP achieves 0.98 with 40 train samples of grasping Fanta (5$\times$ fewer than 200 of DP).}
% \vspace{-0.6cm}
\label{tb:effi}
\end{table}

\section*{Acknowledgements}
This work is supported by National Natural Science Foundation of China under Grants (62361166629, 62225113, 623B2080),
the Major Project of Science and Technology Innovation of Hubei Province (2024BCA003, 2025BEA002), the Innovative Research Group Project of Hubei Province under Grant 2024AFA017, and the Key Research Project of Wuhan City 2024060788020073. The supercomputing system at the Supercomputing Center and the Learning Algorithms \& Soft Manipulation Laboratory of Wuhan University supported the numerical calculations and the robot platforms in this paper.

\clearpage
\bibliography{aaai2026}

\clearpage
\appendix
% \renewcommand\thefigure{A\arabic{figure}}
% \setcounter{figure}{0}
% \renewcommand\thetable{A\arabic{table}}
% \setcounter{table}{0}
% \renewcommand\theequation{A\arabic{equation}}
% \setcounter{equation}{0}
% \pagenumbering{arabic}% resets `page` counter to 1
% \renewcommand*{\thepage}{A\arabic{page}}

% \input{supp}

\end{document}